%% file: conference_101719.tex
\documentclass[conference]{IEEEtran}
\IEEEoverridecommandlockouts
\usepackage{cite}
\usepackage{multirow}%
\usepackage{amsmath,amssymb,amsfonts}
\usepackage{algorithmic}
\usepackage{graphicx}
\usepackage{textcomp}
\usepackage{xcolor}
\usepackage[table]{xcolor}
\usepackage{adjustbox}
\usepackage{booktabs}%
\usepackage{xfp} 
\usepackage[normalem]{ulem} 

\definecolor{shaplow}{RGB}{255,255,204}   
\definecolor{shaphigh}{RGB}{0,150,90}     

\def\BibTeX{{\rm B\kern-.05em{\sc i\kern-.025em b}\kern-.08em
    T\kern-.1667em\lower.7ex\hbox{E}\kern-.125emX}}
\begin{document}

\title{STEP-PD: Stage-Aware and Explainable Parkinson’s Disease Severity Classification Using Multimodal Clinical Assessments}


\author{
\makebox[\textwidth][c]{%
\begin{minipage}{\textwidth}
\centering
\textbf{Md Mezbahul Islam\IEEEauthorrefmark{1}, John Michael Templeton\IEEEauthorrefmark{2}, Christian Poellabauer\IEEEauthorrefmark{1}, Ananda Mohan Mondal\textsuperscript{+}\IEEEauthorrefmark{1}}\\
\vspace{0.3em}
\IEEEauthorrefmark{1}School of Computing \& Information Sciences, Florida International University, Miami, FL, USA\\
\IEEEauthorrefmark{2}College of AI, Cybersecurity and Computing, University of South Florida, Tampa, FL, USA\\
Email: misla093@fiu.edu, jtemplet@usf.edu, cpoellab@fiu.edu, amondal@fiu.edu
\end{minipage}
}
\thanks{\textsuperscript{+}Corresponding author: Ananda Mohan Mondal (amondal@fiu.edu).}%
}



\maketitle

\begin{abstract}
\input{sections/Abstract}
\end{abstract}


\input{sections/Introduction}

\input{sections/Dataset_and_Methodology}

\input{sections/Results}

\input{sections/Discussion}


\bibliographystyle{IEEEtran}
\bibliography{ICHI-bibliography}

\end{document}

%% file: sections/Abstract.tex
Parkinson’s disease (PD) is a progressive disorder in which symptom burden and functional impairment evolve over time, making severity staging essential for clinical monitoring and treatment planning. However, many computational studies emphasize binary PD detection and do not fully exploit repeated follow-up clinical assessments for stage-aware prediction.
This study proposes STEP-PD, a severity-aware machine learning framework to classify PD severity using clinically interpretable boundaries. It leverages all available visits from the Parkinson’s Progression Markers Initiative (PPMI) and integrates routinely collected subjective questionnaires and objective clinician-assessed measures. Disease severity is defined using Hoehn and Yahr staging and grouped into three clinically meaningful categories: Healthy, Mild PD (stages 1--2), and Moderate-to-Severe PD (stages 3--5). Three binary classification problems (Healthy vs. Mild, Healthy vs. Moderate-to-Severe, and Mild vs. Moderate-to-Severe) and a three-class severity task were evaluated using stratified cross-validation with imbalance-aware training. To enhance clinical interpretability, SHAP was employed to provide global explanations (including a cross-task heatmap summarizing stage-dependent symptom relevance) and local, patient-level waterfall explanations.
Across all tasks, XGBoost achieved the strongest and most stable performance, with accuracies of 95.48\% (Healthy vs. Mild), 99.44\% (Healthy vs. Moderate-to-Severe), and 96.78\% (Mild vs. Moderate-to-Severe), and 94.14\% accuracy with 0.8775 Macro-F1 for three-class severity classification. Explainability results highlight a shift from early motor features (e.g., bradykinesia and tremor) to progression-related axial and balance impairments (e.g., postural instability and gait dysfunction).
These findings demonstrate that multimodal clinical assessments within the PPMI cohort can support accurate and interpretable visit-level PD severity stratification.

%% file: sections/Introduction.tex
\section{Introduction}
Parkinson’s disease (PD) is a progressive neurodegenerative disorder that imposes a rapidly growing public health burden and increasing pressure on health-care systems. Recent global projections estimate that PD cases will reach roughly 25 million people worldwide by 2050, largely driven by population aging and demographic change \cite{su2025projections}. In the United States alone, current estimates suggest that approximately 1.1 million people live with PD, and nearly 90,000 new diagnoses occur each year \cite{parkinsonfoundation_stats}. Because PD is heterogeneous and evolves over time, clinicians rely on standardized staging and rating scales (e.g., Hoehn \& Yahr; MDS-UPDRS) to track symptom progression, adjust treatment, and anticipate functional decline—particularly around key milestones where gait and balance impairments emerge and risk of falls increases. These clinical realities motivate machine learning approaches that can support not only PD detection but also \emph{severity-aware} classification and progression-relevant symptom profiling using routinely collected assessments.


Despite rapid progress in machine learning for PD, several practical gaps remain before these models can reliably support severity-aware decision-making in real clinical settings. First, some literature still concentrates on \emph{binary} discrimination (PD vs. Healthy Controls (HC)) and often relies on a single modality (e.g., voice, handwriting, wearables, or imaging), which may limit generalizability and reduce clinical utility when a modality is not available or when symptoms evolve in stages \cite{islam2024review}. Second, although PD is inherently progressive, many studies do not use repeated follow-up assessments from longitudinal cohorts and therefore do not fully exploit progression-relevant variation across visits \cite{gerraty2023machine}. Third, systematic reviews have highlighted a persistent gap between high reported performance and real-world applicability, driven by factors such as heterogeneous validation protocols, limited external testing, and insufficient attention to deployment constraints \cite{tabashum2024machine}. Finally, interpretability is often treated as optional or limited to global feature rankings, although severity staging and treatment planning often require patient-level justification; recent work increasingly calls for explainable frameworks that provide both cohort-level trends and subject-specific reasoning to improve clinician trust and actionability \cite{rabie2025review}.

The motivation for this study stems from two practical needs in PD care and research: (i) moving beyond diagnosis-only models to severity-aware decision support, and (ii) leveraging repeated follow-up assessments available in modern longitudinal cohorts. PPMI is explicitly designed as a multi-center longitudinal observational study to characterize PD progression and support biomarker discovery and clinical trials focused on progression \cite{marek2018parkinson,ppmi_cohorts}. This makes it a natural setting for stage-aware classification using repeated clinical assessments rather than relying only on baseline snapshots. In SCOPE-PD \cite{islam2026scope_pd}, we considered multi-modal (subjective and objective) datasets and binary PD vs. HC classification but did not consider stage-specific changes in measurements and considered only the baseline (first visit). The current study overcomes the limitations and proposes \textbf{STEP-PD}, a machine learning framework that uses routinely collected subjective and objective assessments from all visits for PD severity classification using clinically interpretable boundaries derived from Hoehn and Yahr (H\&Y) staging. Because the emergence of postural instability represents a major functional milestone (typically around H\&Y stage 3) \cite{hy_stages_ncbi}, H\&Y stages were consolidated into three clinically meaningful classes: \textbf{Healthy}, \textbf{Mild/Early PD (Mild)} (stages 1--2), and \textbf{Moderate-to-Severe PD (Mod-Severe)} (stages 3--5). Based on this grouping, three complementary binary classification tasks (Healthy vs. Mild, Healthy vs. Mod--Severe, and Mild vs. Mod--Severe) were evaluated, along with a three-class severity classification setting. This design directly aligns the modeling objective with real-world clinical questions: distinguishing early disease from healthy samples, identifying advanced severity, and characterizing the progression boundary where balance and gait impairment become prominent.

\begin{figure*}[!h]
    \centering    \includegraphics[width=0.95\textwidth]{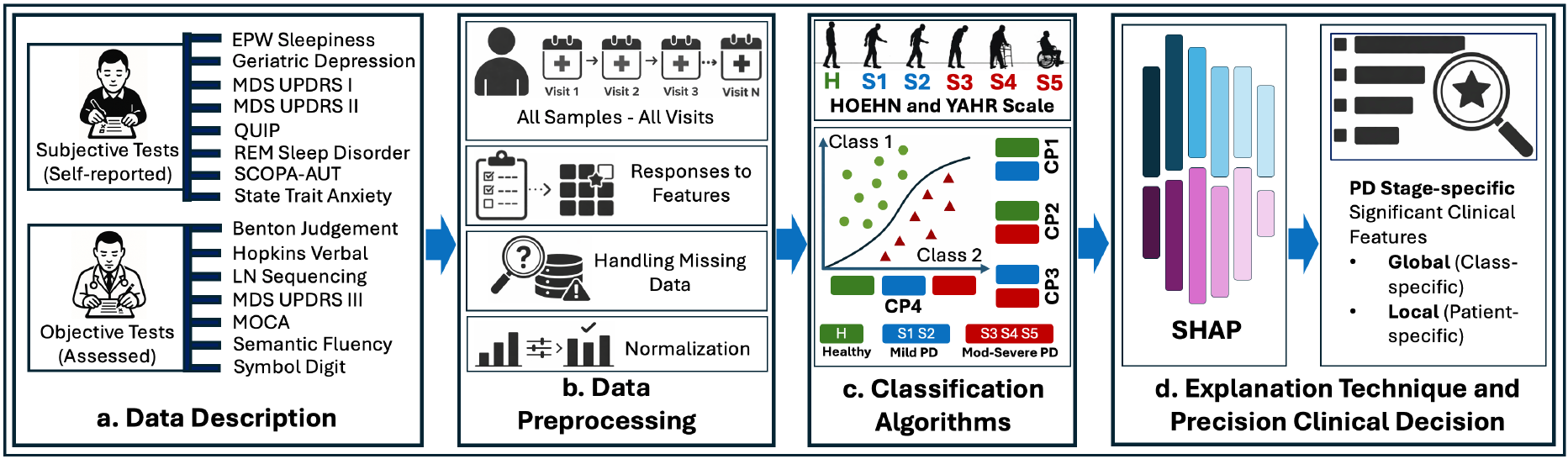}
    \caption{\textbf{Overall Study Framework including Data Description, Data Preprocessing, Classification Algorithms, and Explanation Technique and Precision Clinical Decision.} EPW: Epworth Sleepiness Scale; MDS-UPDRS: Movement Disorder Society Unified Parkinson's Disease Rating Scale;  QUIP: Questionnaire for Impulsive-Compulsive Disorders; REM: Rapid Eye Movement; SCOPA-AUT: Scales for Outcomes in Parkinson’s disease - Autonomic Dysfunction; LN: Letter-Number; MOCA: Montreal Cognitive Assessment; H: Healthy; S: Stage; CP: Classification Problem.
    }
    \label{fig:overall-study-framework}
\end{figure*}

In this study, five widely used supervised machine learning (ML) models were used and evaluated. High predictive accuracy alone is insufficient for clinical translation in PD, where decisions about staging, follow-up, and treatment adjustment benefit from transparent and symptom-based reasoning. Explainable techniques, such as SHapley Additive exPlanations (SHAP)~\cite{lundberg2017unified}, address this issue by clarifying how predictions are made in complex health sectors (e.g., cancer \cite{sobhan2025tilda}, PD~\cite{islam2026scope_pd}). This study also used SHAP as a unified feature attribution framework to provide clinically meaningful explanations.

This study makes the following contributions to clinically interpretable, severity-aware Parkinson’s disease classification using repeated clinical assessments:

\begin{itemize}
    \item \textbf{Severity-aware formulation using multi-visit cohort data}: This study went beyond baseline-only PD detection by leveraging \emph{all available visits} and consolidating H\&Y stages into three clinically meaningful classes (Healthy, Mild, and Mod--Severe), enabling severity evaluation across repeated follow-up assessments.
    
    \item \textbf{Comprehensive benchmarking across clinically aligned tasks:} Five widely used supervised learning models were evaluated across three binary severity problems and one three-class setting using visit-level stratified evaluation with imbalance-aware training.
    
    \item \textbf{Strong predictive performance on stage boundaries:} The proposed framework achieves robust discrimination across all severity settings, including the clinically challenging Mild vs. Mod-Severe boundary, using routinely collected assessments.
    
    \item \textbf{Global and local interpretability for clinical insight:} We provide an explainability layer that combines cohort-level summaries with patient-level rationales, enabling symptom-based interpretation of model predictions.
\end{itemize}

The remainder of this paper is organized as follows. Section~2 describes the PPMI dataset and presents the proposed methodology. Section~3 reports the experimental results, including quantitative performance across classification tasks and interpretability analyses. Section~3 also discusses the main findings, comparisons with previous studies, limitations, and directions for future work.

%% file: sections/Dataset_and_Methodology.tex
\section{Dataset and Methodology}

PD severity is commonly evaluated using standardized clinical rating scales that quantify symptom burden, functional impairment, and disease progression over time. In this study, the disease severity information is derived from the Hoehn and Yahr (H\&Y) \cite{padman2020telediagnosis} staging scale. The Movement Disorder Society–Unified Parkinson’s Disease Rating Scale (MDS-UPDRS) \cite{goetz2008movement} part III provides the H\&Y scale, a globally recognized measure of disease progression. Given its strong measurement properties for motor symptom assessment, high concordance with MDS-UPDRS motor scores, and ability to summarize overall disease status with minimal intra-subject variability, the H\&Y staging scale is used in this work to define disease severity categories for classification.

The H\&Y scale categorizes Parkinson’s disease progression into five stages, ranging from unilateral mild symptoms in Stage 1 to advanced disability requiring wheelchair assistance or continuous care in Stage 5. In this study, stages 1--2 were grouped as \textit{Mild}, and stages 3--5 as \textit{Mod-Severe}, while healthy controls formed the third cohort. Accordingly, the proposed framework considers three groups for classification and explainability: Healthy, Mild, and Mod-Severe. As shown in Figure~\ref{fig:overall-study-framework}, the framework consists of four main components: data description, preprocessing, classification, and explainability-guided clinical interpretation.

\subsection{Data Description}
Data used in this study were obtained from the Parkinson’s Progression Markers Initiative (PPMI) \cite{nalls2015diagnosis}, a large-scale, longitudinal, and publicly available dataset designed to identify biomarkers of PD progression. The study methodology can be found at ppmi-info.org. Data were obtained from the Laboratory of NeuroImaging (LONI). The age range of the participants is 26 to 86 years old. The dataset has multi-modal data, including a comprehensive collection of subjective and objective neuropsychological and clinical assessments. In SCOPE-PD \cite{islam2026scope_pd}, the focus was exclusively on baseline (first-visit) data to distinguish Healthy Controls (HC) from PD patients. While exploring the same dataset for stage information, only stage 1 and stage 2 were found in baseline data, as shown in the ``Stages'' column (the rightmost column) in Table \ref{tab:data_summary}. The absence of samples from later H\&Y stages leads the present study to incorporate all available visits for each participant. Here, all visits are treated as independent visit-level samples. This design allows the framework to leverage repeated follow-up assessments from a longitudinal cohort for severity classification. Samples from all five H\&Y stages were there while considering all visits. The dataset includes eight standardized questionnaires from subjective assessments and seven expert-administered motor and neuropsychological tests from objective assessments as listed in Table \ref{tab:data_summary}. The details of these tests are provided in the prior work \cite{islam2026scope_pd}. A summary of the included assessments, sample counts, and feature composition for both baseline visits and all visits is provided in Table \ref{tab:data_summary}. When restricting the analysis to baseline data, 1,786 participants are present across all tests. When all visits are considered, this expands to 16,162 visit-level samples, substantially increasing the effective sample size for severity-aware modeling. Accordingly, the present study should be interpreted as visit-level severity classification using multi-visit cohort data rather than as a trajectory-based or time-series longitudinal model. Across all assessments, the final feature set comprises 208 features, derived from 230 individual survey questions and clinical items. Detailed descriptions of feature derivation and preprocessing are provided in the Methods section.
\begin{table*}[h]
\caption{\textbf{Summary of Subjective and Objective Tests for baseline and all visits.} CS: Common Samples.}
\label{tab:data_summary}
\centering
\resizebox{0.99\linewidth}{!}{%
\begin{tabular}{@{}l*{8}{c}*{7}{c}@{}cc}
\toprule
\textbf{Test}& \multicolumn{8}{c}{\textbf{Subjective Tests}} & \multicolumn{7}{c}{\textbf{Objective Tests}} & \textbf{Overall}  & \textbf{Stages} \\
\cmidrule(lr){1-1}\cmidrule(lr){2-9} \cmidrule(l){10-16} \cmidrule(lr){17-17} \cmidrule(lr){18-18}
\textbf{Test Name} & 
\rotatebox{90}{EPW Sleepiness \cite{johns1991new}} & \rotatebox{90}{Geriatric Depression \cite{greenberg2012geriatric}}  & \rotatebox{90}{MDS UPDRS I \cite{goetz2008movement}}  & \rotatebox{90}{MDS UPDRS II \cite{goetz2008movement}}  & \rotatebox{90}{QUIP \cite{weintraub2012questionnaire}} & \rotatebox{90}{REM Sleep Disorder \cite{schenck1986chronic}}  & \rotatebox{90}{SCOPA-AUT \cite{visser2004assessment}}  & \rotatebox{90}{State Trait Anxiety\cite{bados2010state}}  
 & 
\rotatebox{90}{Benton Judgement \cite{benton1978visuospatial}} & \rotatebox{90}{Hopkins Verbal \cite{brandt2001hopkins}} & \rotatebox{90}{LN Sequencing \cite{crowe2000does}} & \rotatebox{90}{MDS UPDRS III \cite{goetz2008movement} } & \rotatebox{90}{MOCA \cite{hobson2015montreal}} & \rotatebox{90}{Semantic Fluency \cite{zarino2014new}} & \rotatebox{90}{Symbol Digit \cite{strober2019symbol}} 
 & \rotatebox{90}{\textbf{\# of CS / Total}} & \rotatebox{90}{\textbf{HOEN and YAHR Scale}} \\

\midrule
\textbf{\# of Samples: Baseline Visit} & 
3,832 & 3,761 & 2,212 & 2,212 & 3,829 & 3,839 & 3,830 & 3,760 & 3,813 & 3,819 & 3,813 & 3,817 & 3,813 & 2,562 & 2,370 & \textbf{1,786} & \textbf{1-2}\\
\textbf{\# of Samples: All Visits} & 
18,319 & 18,608 & 31,494 & 31,495 & 18,291 & 18,331
& 18,301 & 18,579 & 16,402 & 16,452 & 16,410 & 34,772 & 17,142 & 16,456 & 16,411 & \textbf{16,162} & \textbf{1-5} \\
\textbf{\# of Survey Questions} & 
8 & 15 & 7 & 13 & 13 & 21 & 21 & 40 & 15 & 7 & 7 & 32 & 27 & 3 & 1 & \textbf{230} & \textbf{--} \\
\textbf{\# of Features} & 
9 & 16 & 7 & 13 & 13 & 20 & 21 & 42  & 1 & 4 & 1 & 32 & 27 & 1 & 1 & \textbf{208} & \textbf{--} \\

\bottomrule
\end{tabular}%
}

\end{table*}
\subsection{Data Preprocessing}
\subsubsection{Converting survey responses into feature values}
Across all assessments and visits, a total of 208 features are extracted from 230 individual clinical items, with continuous scores retained and categorical responses appropriately encoded following PPMI guidelines as shown in Table \ref{tab:data_summary}. This multimodal feature representation enables comprehensive characterization of disease severity and progression at each visit. The conversion process has followed the same process as \cite{islam2026scope_pd} in most cases except for some changes in subjective tests as listed here. The Epworth Sleepiness Scale (EPW) \cite{johns1991new} test includes nine features that come from eight different questions related to sleep and their summation. The Geriatric Depression Scale (GDS) \cite{greenberg2012geriatric} test has 15 questions for patients related to depression, which are expanded 16 features by adding another feature derived from their summation. The REM Sleep Behavior Disorder Questionnaire \cite{schenck1986chronic} test asks 21 questions related to sleep quality and the presence of certain diseases in a patient. One of them asks about the presence of ``PARKISM'' in a patient, which may bias the study for ML-based analysis. We dropped this question and moved on with 20 features from the REM test. The State–Trait Anxiety Inventory (STAI) test includes 40 questions that were kept directly as features, and two features were added by summing up the corresponding response values for state and trait questions. The objective test followed the previous process from \cite{islam2026scope_pd}. The study was completed with all 208 features. 

\subsubsection{Handling missing data}
\begin{figure}[!h]
    \centering    \includegraphics[width=0.99\columnwidth]{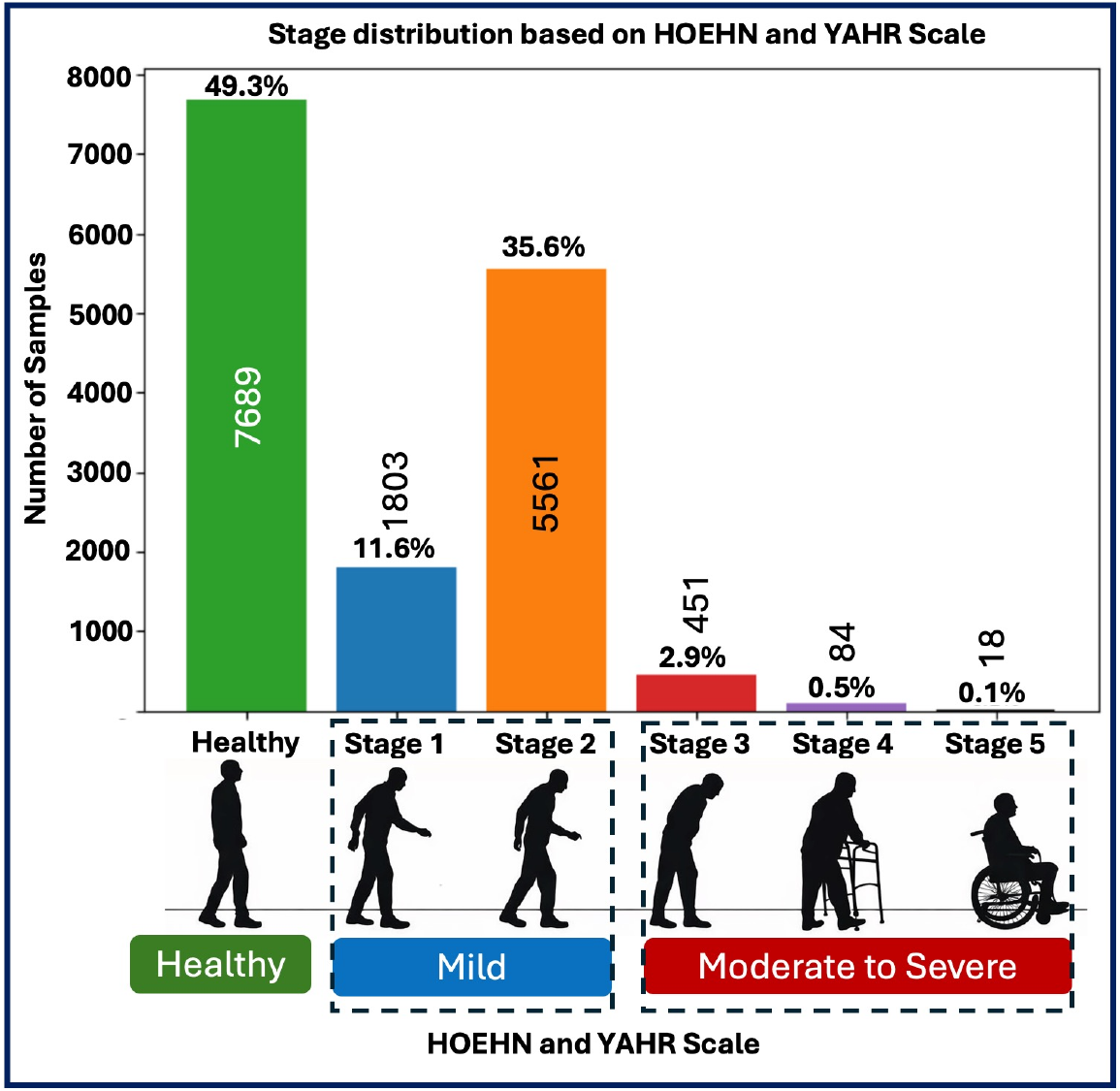}
    \caption{\textbf{Cohort description after data preprocessing.} The mild cohort includes samples from stages 1 and 2, and the Moderate to Severe (Mod-Severe) cohort includes samples from stages 3, 4, and 5. }
    \label{fig:cohort}
\end{figure}

Taking into account all visits as individual samples, the number of samples becomes 16,162. After removing the samples with missing values, it dropped to 15,624 samples. Among the cleaned samples, 7689 are healthy, 1803 are stage 1, 5561 are stage 2, 451 are stage 3, 84 are stage 4, 18 are stage 5, and the other 18 are marked as 101. As the 101 stage information is not clear, they are removed from the final sample list. The distribution of stages in the final 15,606 samples is shown in Figure \ref{fig:cohort}. 

\subsubsection{Normalizing feature values}
The clinical features used in this study originate from heterogeneous assessments with differing ranges of values and statistical properties. For example, items in MDS-UPDRS are scored on ordinal scales (0–4), while several cognitive measures and questionnaire-based measures follow binary or bounded scoring schemes. To ensure comparability between features while preserving relative variability, all numeric features were standardized using z-score normalization.
Specifically, each feature was transformed using the StandardScaler, which rescales the values to have zero mean and unit variance based on the training data distribution. This standardization reduces scale-induced bias, improves numerical stability, and facilitates effective model training, particularly in the presence of features with diverse ranges and variances. The same scaling parameters learned from the training set were subsequently applied to the validation and test data to prevent information leakage. 

\subsection{Classification Algorithms}
\subsubsection{Classification Problems}
To enable clinically interpretable severity-sensitive modeling, the five stages of the H\&Y scale were consolidated into two cohorts: Mild includes Stages 1 and 2, representing unilateral or bilateral symptoms without postural instability and Mod-Severe comprises Stages 3, 4, and 5, characterized by postural instability, gait impairment, and increased functional dependence. Based on this grouping, the final dataset includes 7,689 Healthy samples, 7,364 Mild samples, and 553 Mod-Severe samples. Three binary classification problems (Healthy vs. Mild, Healthy vs. Mod-Severe, and Mild vs. Mod-Severe) and one three-class problem (Healthy vs. Mild vs. Mod-Severe) coming from these three groups will be examined.

\subsubsection{Algorithms}
To benchmark multiple supervised learning strategies for PD severity modeling, we implemented five commonly used classifiers: Logistic Regression (LR), Support Vector Machine (SVM) with a radial basis function kernel, K-Nearest Neighbors (KNN), Random Forest (RF), and Extreme Gradient Boosting (XGBoost). All experiments were conducted in Python using \texttt{scikit-learn} and the \texttt{xgboost} library. Because severity labels derived from H\&Y staging produce an imbalanced class distribution—most notably for the Moderate-to-Severe PD group—class imbalance was handled using class-aware training rather than synthetic oversampling to avoid introducing artificial patterns. For LR, SVM, and RF, we used built-in class balancing (\texttt{class\_weight = "balanced"}) so that each class contributes to the loss in proportion to its inverse frequency. For XGBoost, imbalance was addressed by weighting the minority class using the \texttt{scale\_pos\_weight} parameter \_pos \_weight. In binary settings, this weight was set as:
\begin{equation}
    \text{scale\_pos\_weight} = \frac{n_{\text{major}}}{n_{\text{minor}}}
    \label{eq:xgb_imbalance}
\end{equation}
where $n_{\text{major}}$ and $n_{\text{minor}}$ denote the number of samples in the majority and minority classes, respectively. KNN does not provide a native class-weighting mechanism; therefore, it was evaluated directly under the same stratified sampling protocol.

Hyperparameter tuning and model evaluation followed a 5-fold stratified cross-validation framework. The dataset was first divided into an 80/20 train–test split at the visit level, stratified to preserve class ratios. The training subset (80\%) was used exclusively for model development and hyperparameter optimization through a 5-fold cross-validation (GridSearchCV) procedure, which exhaustively searched predefined parameter grids for each algorithm using the F1-score (two class problem) or macro F1-score (three class problem) as the optimization metric. The hyperparameter configuration with the best performance, identified by the highest mean cross-validated F1-score in all training folds, was then retrained in all 80\% training data and evaluated once in the 20\% test set held.

\subsection{Model Explainability with SHAP}
To improve transparency and clinical interpretability, SHapley Additive exPlanations (SHAP) was applied to quantify how each feature contributes to model predictions at both the individual and cohort levels. For global interpretability, the feature contributions across the cohort were summarized in two complementary ways. First, we computed mean absolute SHAP values separately within each class (e.g., Healthy, Mild, and Mod-Severe) and visualized class-conditional contributions using stacked bar plots. This view identifies the most influential predictors overall and clarifies which class primarily drives each feature’s importance. Second, to compare feature relevance \emph{across tasks} (HC vs. Mild, HC vs. Mod-Severe, and Mild vs. Mod-Severe), a SHAP heatmap was constructed using the union of the top-ranked features. In the heatmap, rows represent features and columns represent classification tasks, with cell intensity reflecting aggregated mean absolute SHAP values. The features can be grouped in seven neurocognitive  functions (NFs) \cite{islam2024redone}. Top features are mapped to the corresponding NF/NFs to get the insight of effected NF. Local explanations were generated using SHAP waterfall (and/or force) plots for representative subjects, illustrating how a patient’s feature values shift the model output from the expected baseline toward the final predicted probability for a given class. These case-level attributions highlight the specific clinical items that provide evidence for or against a severity label and allow clinicians to inspect whether the model decision is driven by plausible symptom patterns.

From a clinical perspective, the heatmap supports decision-making by providing an at-a-glance summary of which symptom domains are most informative at each severity boundary. For example, features that consistently appear with high SHAP values in progression tasks can serve as markers to prioritize during follow-up assessments, while features that are specific to early-stage separation can be useful for screening and early referral. When combined with local waterfall explanations, this framework enables both (i) \textit{population-level insight} into stage-relevant symptom profiles and (ii) \textit{patient-level justification} of individual predictions, facilitating interpretable and clinically grounded severity classification across all settings examined in this study.

%% file: sections/Results.tex
\section{Results and Discussion}

\subsection{t-SNE Visualization of class-specific samples}
\begin{figure*}[!h]
    \centering    \includegraphics[width=0.9\textwidth]{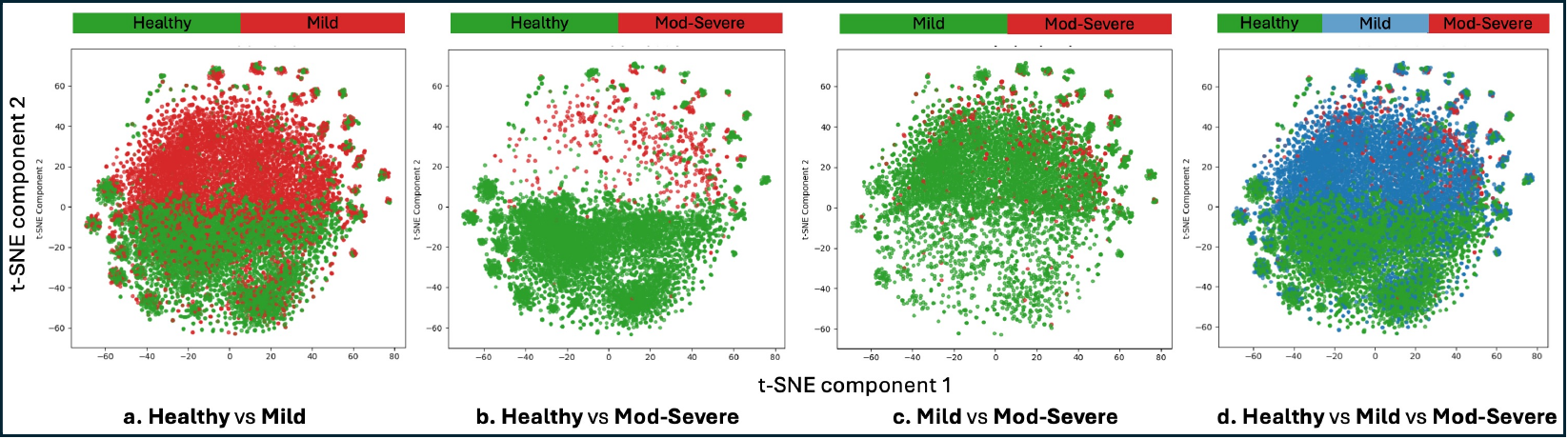}
    \caption{\textbf{t-SNE visualization of the learned feature space for PD stage classification across different diagnostic settings.}
Panels show two-dimensional t-SNE projections of subject representations for (a) Healthy vs. Mild, (b) Healthy vs. Mod–Severe, (c) Mild vs. Mod–Severe, and (d) multi-class classification of Healthy, Mild, and Mod–Severe.
    }
    \label{fig:t-SNE visualization}
\end{figure*}

Figure \ref{fig:t-SNE visualization} illustrates two-dimensional t-SNE projections of the learned feature representations for the four diagnostic settings considered in this study. In the Healthy vs. Mild setting (Figure \ref{fig:t-SNE visualization}.a), the two cohorts exhibit substantial overlap, with only partial separation emerging in localized regions of the embedding space. This pattern reflects the subtle and heterogeneous nature of early Parkinson’s disease, where motor symptoms are present but not yet strongly distinct from healthy aging. The Healthy vs. Mod–Severe comparison (Figure \ref{fig:t-SNE visualization}.b) shows clearer structural separation, with Mod–Severe samples occupying regions that are more distinct from the Healthy cohort. This increased separation is consistent with the presence of more pronounced motor and axial impairments at advanced disease stages. In the Mild vs. Mod–Severe setting (Figure \ref{fig:t-SNE visualization}.c), separation is primarily driven by differences along the dominant embedding directions, although significant overlap remains. This suggests a gradual transition in the learned feature space, consistent with the progressive nature of PD rather than sharply defined stage boundaries. Finally, the three-class visualization (Figure \ref{fig:t-SNE visualization}.d) demonstrates a continuum across Healthy, Mild, and Mod-Severe samples, with Mild largely occupying intermediate regions between the other two groups. While clear class clusters are not strictly isolated, the smooth progression across stages highlights the model’s ability to capture disease severity as a continuous latent structure. Overall, the t-SNE visualizations support the quantitative classification results by revealing stage-dependent structure in the learned feature space while also underscoring the intrinsic overlap and heterogeneity inherent to Parkinson’s disease progression.
\subsection{Experimental Results of Machine Learning Models}
Table~\ref{tab:model_eval_pairs_bold} summarizes the performance of five supervised learning models in three pairwise severity classification tasks and a three-class setting. Overall, tree-based ensembles consistently outperformed distance-based and linear baselines, and XGBoost achieved the best results across all classification problems, showing strong and stable performance over 5-fold cross-validation.

\begin{figure*}[!h]
    \centering    \includegraphics[width=0.9\textwidth]{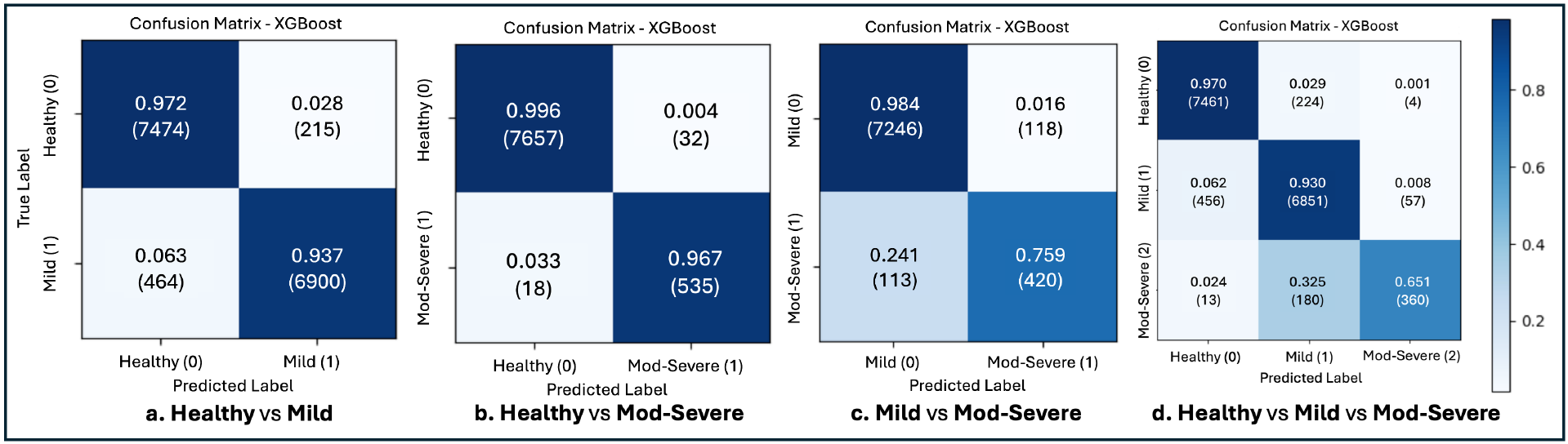}
    \caption{\textbf{Out-of-Fold (OOF) confusion matrix (normalized) for the best ML model.} XGBoost achieved the highest accuracy with all four classification problems. Healthy: Healthy samples; Mild: Stage 1 and Stage 2 PD samples; Mod-Severe (Moderate to severe): Stage 3, Stage 4, and Stage 5 PD samples. 
    }
    \label{fig:confusion-matrix}
\end{figure*}

For Healthy vs. Mild, XGBoost achieved 0.9548 $\pm$ 0.0063 accuracy and 0.9530 $\pm$ 0.0068 F1, with high discrimination (ROC-AUC 0.9888 $\pm$ 0.0022) and strong MCC (0.9099 $\pm$ 0.0124). The corresponding confusion matrix indicates balanced performance for both classes, with most samples correctly assigned to their true label (Figure~\ref{fig:confusion-matrix}.a).

In Healthy vs Mod-Severe, performance increased further due to clearer clinical separation between these groups. XGBoost reached 0.9944 $\pm$ 0.0022 accuracy, 0.9586 $\pm$ 0.0159 F1, and near-ceiling ROC-AUC (0.9983 $\pm$ 0.0022). The confusion matrix shows very few Healthy samples misclassified as Mod-Severe and strong detection of the minority class despite its smaller sample size (Figure~\ref{fig:confusion-matrix}.b), indicating that the model captured stage-relevant patterns beyond simple majority-class prediction.

The most challenging setting was Mild vs. Mod-Severe, where overlap is expected because disease progression is gradual and the Moderate-to-Severe group is substantially smaller. Even in this difficult scenario, XGBoost maintained strong overall performance (accuracy 0.9678 $\pm$ 0.0028, ROC-AUC 0.9775 $\pm$ 0.0077) and achieved the highest F1 (0.7661 $\pm$ 0.0201) and MCC (0.7516 $\pm$ 0.0221) among all models. The confusion matrix reveals that most errors occur when Mod-Severe samples are predicted as Mild (Figure~\ref{fig:confusion-matrix}.c), which is consistent with borderline cases around the transition from H\&Y stage 2 to stage 3. 

Finally, in the three-class classification (Healthy vs. Mild vs. Mod-Severe), XGBoost again performed best with 0.9414 $\pm$ 0.0050 accuracy and 0.8775 $\pm$ 0.0145 Macro-F1, supported by high ROC-AUC (0.9865 $\pm$ 0.0030) and PR-AUC (0.9377 $\pm$ 0.0117). As shown in Figure~\ref{fig:confusion-matrix}.d, Healthy and Mild classes are separated reliably, while the primary confusion remains between Mild and Mod-Severe. This pattern suggests that accuracy alone should be interpreted cautiously in imbalanced severity settings, and that minority-sensitive metrics are essential for evaluating robustness.

Across all settings, XGBoost provides the most consistent and discriminative performance. Importantly, confusion-matrix patterns indicate that misclassifications are largely concentrated near clinically adjacent severity boundaries, which is expected in progressive disorders such as PD. To interpret these performance gains and verify that model decisions are clinically grounded, SHAP-based analysis was used to get feature contributions.

\begin{table*}[t]
\centering
\caption{Performance comparison across classification models for PD stage/severity prediction (mean $\pm$ standard deviation over 5 folds). MCC: Matthews Correlation Coefficient; for the multiclass task (Healthy vs Mild vs Mod-Severe), F1 denotes Macro-F1; Mod-Severe: Moderate to Severe.}
\label{tab:model_eval_pairs_bold}
\resizebox{\linewidth}{!}{
\begin{tabular}{llccccc}
\toprule
\textbf{Classification Problem} & \textbf{Model} & \textbf{Accuracy $\pm$ SD} & \textbf{F1 $\pm$ SD} & \textbf{ROC AUC $\pm$ SD} & \textbf{PR AUC $\pm$ SD} & \textbf{MCC $\pm$ SD} \\
\midrule

\multirow{5}{*}{\textbf{Healthy vs. Mild}} 
 & KNN                 & 0.8497 $\pm$ 0.0037 & 0.8263 $\pm$ 0.0042 & 0.9143 $\pm$ 0.0109 & 0.9098 $\pm$ 0.0187 & 0.7168 $\pm$ 0.0082 \\
 & Logistic Regression & 0.9422 $\pm$ 0.0036 & 0.9392 $\pm$ 0.0039 & 0.9746 $\pm$ 0.0037 & 0.9714 $\pm$ 0.0093 & 0.8855 $\pm$ 0.0071 \\
 & Random Forest       & 0.9546 $\pm$ 0.0063 & 0.9528 $\pm$ 0.0066 & 0.9880 $\pm$ 0.0024 & 0.9881 $\pm$ 0.0031 & 0.9094 $\pm$ 0.0125 \\
 & SVM (RBF)           & 0.9490 $\pm$ 0.0071 & 0.9464 $\pm$ 0.0078 & 0.9827 $\pm$ 0.0032 & 0.9840 $\pm$ 0.0038 & 0.8993 $\pm$ 0.0140 \\
 & \textbf{XGBoost}            & \textbf{0.9548 $\pm$ 0.0063} & \textbf{0.9530 $\pm$ 0.0068} & \textbf{0.9888 $\pm$ 0.0022} & \textbf{0.9892 $\pm$ 0.0024} & \textbf{0.9099 $\pm$ 0.0124} \\
\midrule

\multirow{5}{*}{\textbf{Healthy vs. Mod-Severe}} 
 & KNN                 & 0.9749 $\pm$ 0.0032 & 0.7781 $\pm$ 0.0322 & 0.9205 $\pm$ 0.0130 & 0.8066 $\pm$ 0.0361 & 0.7804 $\pm$ 0.0316 \\
 & Logistic Regression & 0.9823 $\pm$ 0.0030 & 0.8785 $\pm$ 0.0191 & 0.9891 $\pm$ 0.0063 & 0.9209 $\pm$ 0.0387 & 0.8724 $\pm$ 0.0196 \\
 & Random Forest       & 0.9937 $\pm$ 0.0025 & 0.9514 $\pm$ 0.0195 & 0.9979 $\pm$ 0.0024 & 0.9890 $\pm$ 0.0054 & 0.9486 $\pm$ 0.0205 \\
 & SVM (RBF)           & 0.9890 $\pm$ 0.0026 & 0.9174 $\pm$ 0.0184 & 0.9912 $\pm$ 0.0025 & 0.9559 $\pm$ 0.0155 & 0.9121 $\pm$ 0.0194 \\
 & \textbf{XGBoost}             & \textbf{0.9944 $\pm$ 0.0022} & \textbf{0.9586 $\pm$ 0.0159} & \textbf{0.9983 $\pm$ 0.0022} & \textbf{0.9908 $\pm$ 0.0055} & \textbf{0.9556 $\pm$ 0.0170} \\
\midrule

\multirow{5}{*}{\textbf{Mild vs. Mod-Severe}} 
 & KNN                 & 0.9376 $\pm$ 0.0067 & 0.3005 $\pm$ 0.0900 & 0.7094 $\pm$ 0.0205 & 0.2977 $\pm$ 0.0710 & 0.3404 $\pm$ 0.0999 \\
 & Logistic Regression & 0.9027 $\pm$ 0.0078 & 0.5349 $\pm$ 0.0355 & 0.9272 $\pm$ 0.0263 & 0.5812 $\pm$ 0.0336 & 0.5237 $\pm$ 0.0426 \\
 & Random Forest       & 0.9574 $\pm$ 0.0027 & 0.6692 $\pm$ 0.0143 & 0.9699 $\pm$ 0.0092 & 0.7454 $\pm$ 0.0237 & 0.6506 $\pm$ 0.0161 \\
 & SVM (RBF)           & 0.9370 $\pm$ 0.0049 & 0.5737 $\pm$ 0.0336 & 0.9195 $\pm$ 0.0220 & 0.5992 $\pm$ 0.0270 & 0.5409 $\pm$ 0.0361 \\
 & \textbf{XGBoost}             & \textbf{0.9678 $\pm$ 0.0028} & \textbf{0.7661 $\pm$ 0.0201} & \textbf{0.9775 $\pm$ 0.0077} & \textbf{0.8472 $\pm$ 0.0276} & \textbf{0.7516 $\pm$ 0.0221} \\
\midrule

\multirow{5}{*}{\textbf{Healthy vs. Mild vs. Mod-Severe}} 
 & KNN                 & 0.8221 $\pm$ 0.0052 & 0.6520 $\pm$ 0.0230 & 0.8434 $\pm$ 0.0112 & 0.6816 $\pm$ 0.0141 & 0.6721 $\pm$ 0.0104 \\
 & Logistic Regression & 0.8803 $\pm$ 0.0074 & 0.7557 $\pm$ 0.0122 & 0.9530 $\pm$ 0.0038 & 0.8043 $\pm$ 0.0181 & 0.7894 $\pm$ 0.0119 \\
 & Random Forest       & 0.9328 $\pm$ 0.0067 & 0.8543 $\pm$ 0.0158 & 0.9835 $\pm$ 0.0027 & 0.9004 $\pm$ 0.0125 & 0.8738 $\pm$ 0.0127 \\
 & SVM (RBF)           & 0.9223 $\pm$ 0.0052 & 0.8113 $\pm$ 0.0104 & 0.9680 $\pm$ 0.0042 & 0.8461 $\pm$ 0.0129 & 0.8551 $\pm$ 0.0100 \\
 & \textbf{XGBoost}            & \textbf{0.9414 $\pm$ 0.0050} & \textbf{0.8775 $\pm$ 0.0145} & \textbf{0.9865 $\pm$ 0.0030} & \textbf{0.9377 $\pm$ 0.0117} & \textbf{0.8898 $\pm$ 0.0095} \\
\bottomrule
\end{tabular}
}
\end{table*}

\subsection{Cohort-Specific Significant Clinical Features}
\begin{figure*}[!h]
    \centering    \includegraphics[width=0.99\textwidth]{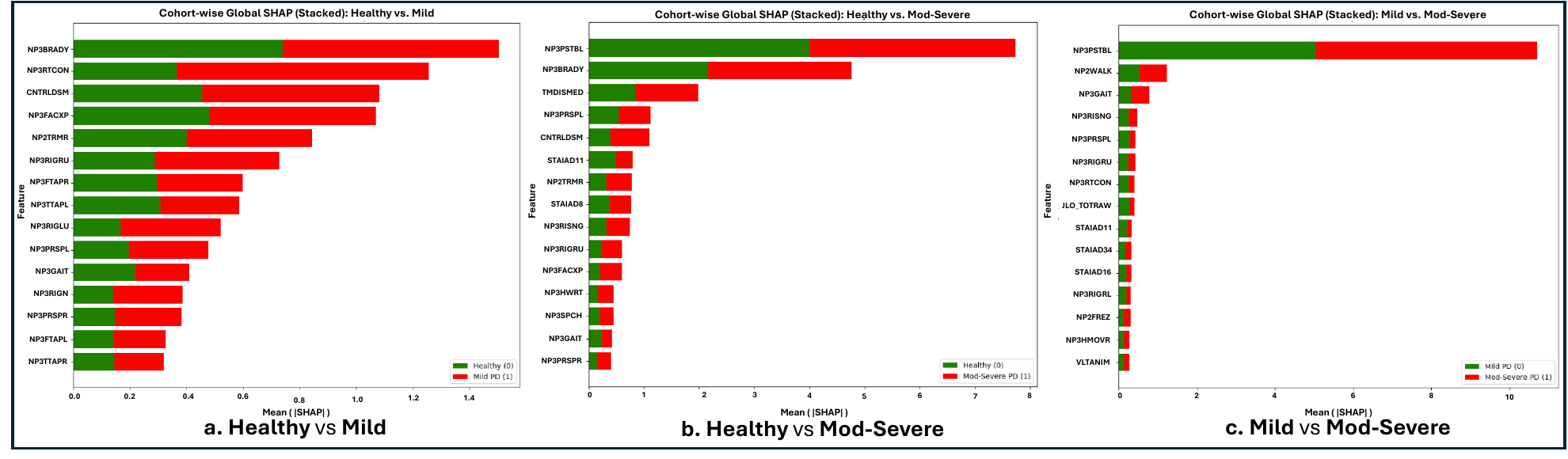}
    \caption{\textbf{Cohort-wise global feature contributions for three Parkinson’s disease classification tasks.}
Stacked bar plots show the mean absolute SHAP values of the 15 top features, decomposed by cohort contribution, for (a) Healthy vs. Mild, (b) Healthy vs. Mod-Severe, and (c) Mild vs. Mod-Severe. Green bars indicate feature contributions to the less severe cohort, while red bars represent feature contributions to the more severe cohort. 
}
    \label{fig:global_shap}
\end{figure*}

Figure \ref{fig:global_shap} presents the cohort-wise global top 15 feature contributions derived from SHAP analysis across the three pairwise Parkinson’s disease classification tasks. The description of each feature is provided in table \ref{tab:shap_heatmap_union_all}. The stacked representation allows direct comparison of how individual features contribute differently to each cohort within a given task.
In the Healthy vs. Mild comparison (Figure \ref{fig:global_shap}.a), global importance is dominated by classical early motor features, including bradykinesia, rigidity, tremor, and facial expressivity. Contributions are relatively balanced between cohorts, reflecting subtle but discriminative motor changes that emerge during early disease onset.
For the Healthy vs. Mod-Severe task (Figure \ref{fig:global_shap}.b), the contribution pattern shifts markedly toward features associated with advanced motor impairment. Postural instability and bradykinesia emerge as the most influential features, with substantially larger contributions from the Mod-Severe PD cohort. Additional importance from medication-related variables and speech measures suggests increased clinical complexity and treatment effects at later stages.
The Mild PD vs. Mod-Severe PD comparison (Figure \ref{fig:global_shap}.c) is overwhelmingly driven by postural instability and gait-related features, including walking difficulty, rising from a chair, and freezing of gait. Here, contributions are strongly skewed toward the Mod-Severe PD cohort, highlighting axial and balance impairments as key markers of disease progression beyond the mild stage.
Overall, these results demonstrate a coherent progression in global feature importance, transitioning from dopaminergic-responsive motor symptoms in early disease to axial, gait, and postural dysfunction in advanced Parkinson’s disease.

\subsection{Symptoms Associated with Disease Progression}

To summarize and compare stage-specific drivers across the three binary classification settings, we constructed the SHAP heatmap shown in Table~\ref{tab:shap_heatmap_union_all}. For each binary task (Healthy vs. Mild, Healthy vs. Mod–Severe, and Mild vs. Mod–Severe), we first selected the top 15 features ranked by total mean absolute SHAP score. We then took the union of these three sets, resulting in 30 unique features, and visualized their cohort-level contributions across tasks. Each cell represents the mean of absolute SHAP values for the combined cohort for a given feature–task pair, where darker shading (green) indicates a stronger global influence on model decisions within that task. In addition to reporting each feature’s originating test and clinical domain, the table includes a Neurocognitive Function (NF) label that maps features to broader functional categories (e.g., Motor, Behavioral/Psychological, Executive Function, Memory, and Speech), enabling a clinically interpretable view of which functional systems drive separation at different severity boundaries.

In the heatmap of the healthy vs. Mild task, high-importance features concentrates around classic early motor signatures (NF: Motor), including global bradykinesia (NP3BRADY), tremor- and rigidity-related items, and fine motor control measures such as finger and toe tapping. In contrast, the Healthy vs Mod–Severe and Mild vs Mod–Severe tasks are dominated by axial and balance-related impairments (NF: Motor), most notably postural instability (NP3PSTBL), along with measures of gait and arising from the chair. Beyond motor examination features, the heatmap also highlights the contribution of non-motor and behavioral measures. Anxiety-related items (STAIAD*) and medication-related behavior variables (QUIP-derived features) are labeled under NF: Behavioral/Psychological and appear primarily in comparisons involving more severe disease. It suggests that non-motor burden and treatment-related complications provide additional discriminative signals at later stages. Cognitive measures (e.g., JLO\_TOTRAW and VLTANIM), tagged under NF categories such as Executive Function, Memory, and Speech, emerge in the Mild vs. Mod–Severe comparison, indicating that cognitive and visuospatial changes can also contribute to distinguishing more advanced disease in this cohort.

From a clinical perspective, this heatmap is useful because it does more than rank features and clarifies which symptom domains and broader neurocognitive functions matter most at each diagnostic boundary. As a result, it can help interpret model behavior in a stage-aware way: Motor-dominant features emphasized in early-stage discrimination may support screening and early referral, while Motor and balance-related features that rise in importance in progression tasks (e.g., postural instability and gait dysfunction) align with clinical priorities during follow-up monitoring and risk assessment. When combined with individual-level SHAP waterfall explanations, the heatmap supports both population-level insight and sample-specific interpretation of the functional drivers underlying each prediction, as explained in the next section.

\begin{table*}[t]
\centering
\caption{Heatmap of mean absolute SHAP values for the combined cohort (e.g., Healthy + Mild in Healthy vs. Mild) across three PD classification tasks. Cell shading uses a yellow-to-green scale (yellow = lower importance, green = higher importance) within each task column. Feature importance shifts from bradykinesia- and rigidity-related motor signs in early disease to postural instability, gait dysfunction, and axial symptoms in later disease stages. \textbf{H}: Healthy; \textbf{NF}: Neurocognitive Function; \textbf{B/P}: Behavior/Psychological; \textbf{E\_F}: Executive Function; \textbf{Sp}: Speech.}

\label{tab:shap_heatmap_union_all}
\scriptsize
\setlength{\tabcolsep}{4pt}
\renewcommand{\arraystretch}{1.15}
\adjustbox{width=\textwidth}{
\begin{tabular}{lllllp{4cm}ccc}
\toprule
\textbf{Feature} &
\textbf{Test} &
\textbf{S/O} &
\textbf{Domain} &
\textbf{NF/NFs} &
\textbf{Feature Description} &
\textbf{H vs Mild} &
\textbf{H vs Mod--Sev} &
\textbf{Mild vs Mod--Sev} \\
\midrule

NP3PSTBL & MDS UPDRS III & O & Postural Stability & Motor & Postural stability (pull test / balance) &
 & \cellcolor{shaphigh!100!shaplow}7.74 & \cellcolor{shaphigh!100!shaplow}10.71 \\

NP3BRADY & MDS UPDRS III & O & Bradykinesia & Motor & Global bradykinesia score &
\cellcolor{shaphigh!100!shaplow}1.50 & \cellcolor{shaphigh!67!shaplow}4.76 &  \\

TMDISMED & QUIP & S & Medication Effect & B/P & Consistently take too much of Parkinson’s medications &
 & \cellcolor{shaphigh!37!shaplow}1.98 &  \\

NP3RTCON & MDS UPDRS III & O & Tremor & Motor & Constancy of rest tremor when different body parts are variously at rest &
\cellcolor{shaphigh!86!shaplow}1.26 &  & \cellcolor{shaphigh!18!shaplow}0.41 \\

NP2WALK & MDS UPDRS II & S & Gait & Motor & Walking and balance during daily activities &
 &  & \cellcolor{shaphigh!25!shaplow}1.24 \\

NP3PRSPL & MDS UPDRS III & O & Pronation--Supination & Motor & Pronation--supination (left hand) &
\cellcolor{shaphigh!42!shaplow}0.48 & \cellcolor{shaphigh!27!shaplow}1.13 & \cellcolor{shaphigh!19!shaplow}0.44 \\

CNTRLDSM & QUIP & S & Behavior & B/P & Difficulty in controlling the use of Parkinson’s medications &
\cellcolor{shaphigh!76!shaplow}1.08 & \cellcolor{shaphigh!27!shaplow}1.10 &  \\

NP3FACXP & MDS UPDRS III & O & Facial Expression & Motor & Facial expression (hypomimia) &
\cellcolor{shaphigh!75!shaplow}1.07 & \cellcolor{shaphigh!22!shaplow}0.60 &  \\

NP2TRMR & MDS UPDRS II & S & Tremor & Motor & Tremor during daily activities &
\cellcolor{shaphigh!63!shaplow}0.84 & \cellcolor{shaphigh!24!shaplow}0.78 &  \\

STAIAD11 & State Trait Anxiety & S & Anxiety & B/P & Issues regarding self-confident &
 & \cellcolor{shaphigh!24!shaplow}0.80 & \cellcolor{shaphigh!18!shaplow}0.34 \\

STAIAD8 & State Trait Anxiety & S & Anxiety & B/P & Issues regarding satisfaction &
 & \cellcolor{shaphigh!24!shaplow}0.77 &  \\

NP3RISNG & MDS UPDRS III & O & Motor & Motor & Arising from chair &
 & \cellcolor{shaphigh!23!shaplow}0.74 & \cellcolor{shaphigh!19!shaplow}0.47 \\

NP3RIGRU & MDS UPDRS III & O & Rigidity & Motor & Rigidity (right upper limb) &
\cellcolor{shaphigh!56!shaplow}0.73 & \cellcolor{shaphigh!22!shaplow}0.60 & \cellcolor{shaphigh!18!shaplow}0.43 \\

NP3RIGLU & MDS UPDRS III & O & Rigidity & Motor & Rigidity (left upper limb) &
\cellcolor{shaphigh!45!shaplow}0.52 &  &  \\

NP3GAIT & MDS UPDRS III & O & Gait & Motor & Gait (walking performance) &
\cellcolor{shaphigh!38!shaplow}0.41 & \cellcolor{shaphigh!20!shaplow}0.42 & \cellcolor{shaphigh!21!shaplow}0.80 \\

NP3RIGN & MDS UPDRS III & O & Rigidity & Motor & Rigidity (neck) &
\cellcolor{shaphigh!37!shaplow}0.39 &  &  \\

NP3PRSPR & MDS UPDRS III & O & Pronation--Supination & Motor & Pronation--supination (right hand) &
\cellcolor{shaphigh!37!shaplow}0.38 & \cellcolor{shaphigh!20!shaplow}0.41 &  \\

NP3FTAPL & MDS UPDRS III & O & Finger Tapping & Motor & Finger tapping (left hand) &
\cellcolor{shaphigh!33!shaplow}0.33 &  &  \\

NP3TTAPR & MDS UPDRS III & O & Toe Tapping & Motor & Toe tapping (right foot) &
\cellcolor{shaphigh!33!shaplow}0.32 &  &  \\

NP3FTAPR & MDS UPDRS III & O & Finger Tapping & Motor & Finger tapping (right hand) &
\cellcolor{shaphigh!49!shaplow}0.60 &  &  \\

NP3TTAPL & MDS UPDRS III & O & Toe Tapping & Motor & Toe tapping (left foot) &
\cellcolor{shaphigh!48!shaplow}0.59 &  &  \\

NP2HWRT & MDS UPDRS II & S & Handwriting & Motor & Handwriting impairment &
 & \cellcolor{shaphigh!20!shaplow}0.46 &  \\

NP2SPCH & MDS UPDRS II & S & Speech & Sp & Speech difficulties in daily life &
 & \cellcolor{shaphigh!20!shaplow}0.45 &  \\

NP3RIGRL & MDS UPDRS III & O & Rigidity & Motor & Rigidity (right lower limb) &
 &  & \cellcolor{shaphigh!17!shaplow}0.31 \\

NP2FREZ & MDS UPDRS II & S & Freezing of Gait & Motor & Freezing episodes during walking &
 &  & \cellcolor{shaphigh!17!shaplow}0.31 \\

NP3HMOVR & MDS UPDRS III & O & Hand Movements & Motor & Hand movements (right hand) &
 &  & \cellcolor{shaphigh!17!shaplow}0.28 \\

JLO\_TOTRAW & Benton Judgement & O & Cognition & E\_F & Visuospatial judgment total raw score &
 &  & \cellcolor{shaphigh!18!shaplow}0.41 \\

VLTANIM & Semantic Fluency & O & Cognition & E\_F, Sp & Animal naming (semantic fluency) &
 &  & \cellcolor{shaphigh!17!shaplow}0.28 \\

STAIAD34 & State Trait Anxiety & S & Anxiety & B/P & Issues regarding making decisions easily &
 &  & \cellcolor{shaphigh!18!shaplow}0.34 \\

STAIAD16 & State Trait Anxiety & S & Anxiety & B/P & Issues regarding feeling content &
 &  & \cellcolor{shaphigh!18!shaplow}0.33 \\

\bottomrule
\end{tabular}
}
\end{table*}

\subsection{Sample-Specific Significant Clinical Features}
\begin{figure}[!h]
    \centering    \includegraphics[width=0.99\columnwidth]{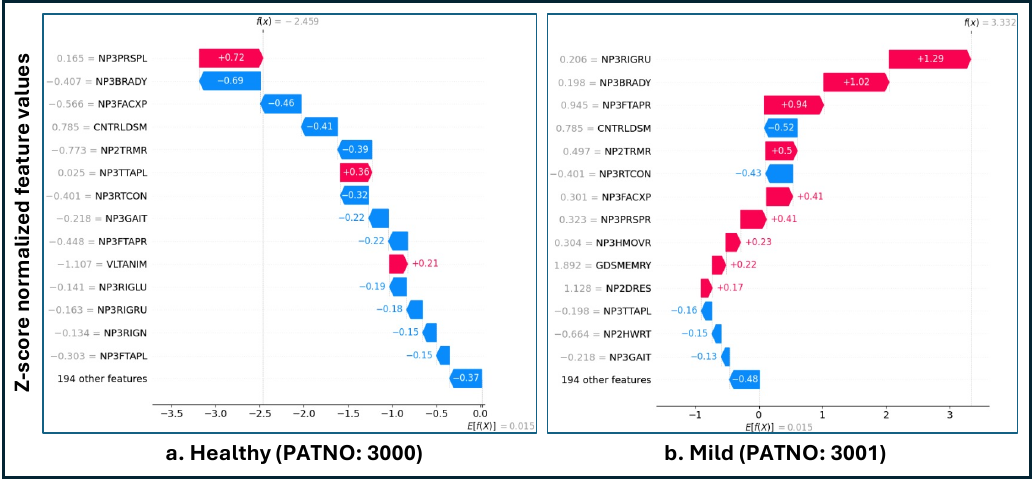}
    \caption{\textbf{Sample-specific feature contributions for a Healthy (patient No: 3000) and Mild (patient No: 3001) sample.} }
    \label{fig:local_shap}
\end{figure}

To complement the global interpretability analysis, we further examine local feature attributions using SHAP waterfall plots for individual predictions. Figure \ref{fig:local_shap} presents representative examples of the Healthy vs. Mild PD classification task, including one Healthy subject (Figure \ref{fig:local_shap}.a) and one Mild subject (Figure \ref{fig:local_shap}.b).
For the Healthy subject, the prediction is primarily driven by negative contributions from motor features associated with bradykinesia, rigidity, and gait, which collectively shift the model output toward the Healthy class. In contrast, for the Mild PD subject, positive contributions from bradykinesia, rigidity, and coordination-related features increase the predicted probability of Mild PD as expected, while a smaller number of features provide opposing evidence.
These examples demonstrate how the proposed model produces clinically interpretable subject-specific explanations. Overall, this local interpretability capability supports transparent model decision-making and highlights the potential for personalized clinical insight by linking individual predictions directly to the underlying symptom profiles.

%% file: sections/Discussion.tex
\subsection{Discussion}
\textbf{Objective and key contributions:}  
The goal of this work is to go beyond binary detection of Parkinson's disease and enable interpretable \emph{severity-aware} classification using clinical evaluations from a longitudinal cohort. Specifically, we targeted three clinically meaningful classification problems derived from H\&Y staging: (i) Healthy vs. Mild, (ii) Healthy vs. Mod-Severe, and (iii) Mild vs. Mod-Severe. The classification tasks were complemented with global and local explainability through SHAP. At the modeling level, we evaluated multiple supervised algorithms and found that ensemble methods—particularly \textbf{XGBoost}—provided the most consistent and discriminative performance across all tasks (Table~\ref{tab:model_eval_pairs_bold}). 

\textbf{Performance trends across severity settings:}  
The results show a coherent progression in task difficulty. Separation of Healthy vs. Mod-Severe achieved near-ceiling performance, which is expected given the stronger symptom burden and clearer functional impairment in advanced disease. Distinguishing \textbf{Healthy vs. Mild} remained highly accurate, indicating that the multimodal clinical feature set captures subtle early-stage motor and non-motor differences. The most challenging setting was \textbf{Mild vs. Mod-Severe} (low F1 and MCC score in Table \ref{tab:model_eval_pairs_bold}), where symptoms overlap and the boundary between H\&Y stage 2 and stage 3 represents a gradual clinical transition. The confusion-matrix patterns are consistent with this interpretation, and most errors occur between adjacent severity categories rather than between extremes (Figure \ref{fig:confusion-matrix}).

\textbf{Clinical interpretability via SHAP:}  
Explainability analyses further connect model behavior to clinical expectations. Global SHAP summaries and the cross-task heatmap (Table~\ref{tab:shap_heatmap_union_all}) reveal stage-dependent changes in feature importance: early-stage discrimination is primarily driven by classic motor findings such as bradykinesia, rigidity, tremor, and fine motor control, while later-stage comparisons emphasize axial and balance-related impairments, particularly postural instability and gait dysfunction. In addition, non-motor and behavioral signals (e.g., anxiety items, cognition, and medication-related behaviors) emerge more prominently in comparisons involving advanced disease, reflecting increased clinical complexity at later stages. Local SHAP waterfall plots provide case-level justification by highlighting the specific symptoms that push an individual prediction toward a severity label, which supports transparent clinical review and facilitates exploration of symptom profiles across samples and tasks. At the same time, because several influential features are derived from MDS-UPDRS items that are clinically related to H\&Y staging, these explanations should be interpreted in light of potential target-feature overlap.

\textbf{Comparison with existing literature:}  
As summarized in Table~\ref{tab:literature_compare}, previous PD classification studies often rely on a single modality—such as speech, wearable sensors, or imaging—or focus primarily on binary PD vs. HC detection. In contrast, the proposed \textbf{STEP-PD} framework leverages multimodal clinical assessments and explicitly targets severity-aware classification. Instead of generating synthetic samples, imbalance was handled through class-aware learning and stratified evaluation, which reduces the risk of introducing artificial patterns in minority severity classes. Importantly, the accuracies achieved for Healthy vs. Mild (95.48\%), Healthy vs. Mod-Severe (99.44\%), and Mild vs. Mod-Severe (96.78\%) demonstrate competitive performance relative to existing approaches while offering the added advantage of clinical transparency through SHAP-based explanations. Broader generalizability remains to be established through subject-aware and external validation.

\begin{table}[ht]
\centering
\caption{Comparison of Parkinson's disease classification studies and the proposed STEP-PD models.}
\label{tab:literature_compare}
\resizebox{\columnwidth}{!}{
\begin{tabular}{@{}p{2.5cm} p{3.5cm} p{1.2cm} p{1.5cm} p{3cm}@{}}
\toprule
\textbf{Author} & \textbf{Data} & \textbf{Accuracy} & \textbf{Dataset} & \textbf{Classification} \\
\cmidrule(lr){1-1} \cmidrule(lr){2-5}

Grover et al.~\cite{grover2018predicting} & Voice recordings & 81.67\% & UCI ML Repository & Severe vs. Not Severe \\
Templeton et al.~\cite{templeton2022classification} & Sensor-based metric & 92.60\% & Self-Collected & PD vs. HC \\
Adebimpe et al.~\cite{esan2025explainable} & Subjective (UPDRS, MoCA) & 93\% & PPMI & PD vs. HC \\
Dentamaro et al.~\cite{dentamaro2024enhancing} & 3D MRI Image, Clinical & 96.60\% & PPMI & Prodromal stages \\
Priyadharshini et al.~\cite{priyadharshini2024comprehensive} & T2-weighted 3D MRI & 96.80\% & PPMI & PD vs. Prodromal vs. HC \\
SCOPE-PD & Subjective and Objective measurements (baseline visits) & 98.60\% & PPMI & PD vs. HC \\
\midrule
\textbf{STEP-PD: Healthy vs. Mild} & Subjective and Objective measurements (all-visits) & 95.48\% & PPMI & Mild PD vs. HC \\

\textbf{STEP-PD: Healthy vs. Mod-Severe} & Subjective and Objective measurements (all-visits) & \textbf{99.44}\% & PPMI & Moderate to Severe PD vs. HC \\

\textbf{STEP-PD: Mild vs. Mod-Severe} & Subjective and Objective measurements (all-visits) & 96.78\% & PPMI & Mild PD vs. Moderate to Severe PD\\
\bottomrule
\end{tabular}
}
\footnotesize
\end{table}

\textbf{Limitations and future directions.}  
This study has several limitations. First, because H\&Y staging is clinically related to motor symptom severity, and several influential predictors are derived from MDS-UPDRS items, part of the observed performance may reflect reconstruction of staging-related clinical structure rather than discovery of fully independent multimodal severity signals. Second, although the dataset is longitudinal at the cohort level, the current framework performs visit-level classification and does not explicitly model temporal dependencies, subject-specific trajectories, or visit-to-visit transitions. Third, because evaluation was performed at the visit level rather than the subject level, repeated visits from the same participant may appear across data partitions, which can introduce within-subject dependency and may yield optimistic performance estimates relative to stricter subject-aware grouped validation. Fourth, the Mod-Severe cohort remains comparatively small, with especially limited representation of stages 4 and 5, so minority-class recognition remains more challenging than overall accuracy alone may suggest. Finally, the current analysis is limited to the PPMI cohort, and external validation in independent cohorts is needed before broader claims regarding real-world clinical applicability can be made. Overall, STEP-PD demonstrates that repeated clinical assessments from a longitudinal cohort can support accurate and interpretable PD severity stratification within PPMI, while subject-aware evaluation and external validation remain important next steps.

\textbf{Data Acknowledgment}: Data used in the preparation of this article were obtained [on May 3, 2025] from the Parkinson’s Progression Markers Initiative (PPMI) database (https://www.ppmi-info.org/access-data-specimens/download-data), RRID:SCR 006431. For up-to-date information on the study, visit http://www.ppmi-info.org. PPMI – a public-private partnership – is funded by the Michael J. Fox Foundation for Parkinson’s Research and funding partners, including 4D Pharma, Abbvie, et al..

\textbf{Funding}: This work was partially supported by the NIH/NCI R21CA290324 and NIH/NHGRI UG3HG013615. The content is solely the responsibility of the authors and does not necessarily represent the official views of the funding agencies.

\textbf{Conflict of Interest Statement}: All authors of this work declare that there are no conflicts of interest in the authorship nor publication of this contribution.